\documentclass{article}

\usepackage[preprint]{colm2026_conference}

\usepackage[T2A,T1]{fontenc}
\usepackage[utf8]{inputenc}
\usepackage[english]{babel}
\usepackage{microtype}
\usepackage{hyperref}
\usepackage{url}
\usepackage{xcolor}
\usepackage{graphicx}
\usepackage{amsmath}
\usepackage{algorithm}
\usepackage{algorithmic}
\usepackage{booktabs}
\usepackage{caption}
\usepackage{placeins}
\usepackage{rotating}
\usepackage{tikz}
\usepackage{lineno}
\usepackage[scaled=0.85]{DejaVuSansMono}
\usetikzlibrary{arrows.meta,positioning,shapes.geometric,calc}

\definecolor{darkblue}{rgb}{0, 0, 0.5}
\hypersetup{
  colorlinks=true,
  citecolor=darkblue,
  linkcolor=darkblue,
  urlcolor=darkblue,
  pdftitle={Writing-System-Level Tokenizer Adaptation for Byte-Level BPE},
  pdfauthor={Bohdan Didenko}
}

\newcommand{\cy}[1]{{\fontencoding{T2A}\selectfont #1}}

\title{Writing-System-Level Tokenizer Adaptation for Byte-Level BPE}

\author{Bohdan Didenko\\
\normalfont Lviv Polytechnic National University}

\begin{document}
\maketitle
\lhead{Preprint}

\begin{abstract}
Pretrained byte-level BPE tokenizers can segment underrepresented languages inefficiently.
Replacing a tokenizer changes the meaning of nearly every token ID, while vocabulary expansion enlarges the model's embedding and output matrices.
We study post-hoc adaptation that keeps the model-vocabulary size fixed and preserves most existing token-to-ID assignments as a construction-time compatibility property.
Directly transferring tokens from a language-specific tokenizer does not guarantee derivability through the target BPE merge graph: an inserted entry can conflict with the target's greedy merge ranks.
We formalize this failure as the \textbf{merge ordering problem} and introduce \textbf{BPE-guided insertion}, which builds each transferred token through a target-reachable decomposition.
Our pipeline uses script-aware row selection to limit collateral fragmentation, reconstructs target-script byte-level prerequisites, and applies guided insertion to maintain merge-graph reachability.
On Ukrainian adaptations of Nemotron and GPT-OSS, it reduces token counts by 33.5\% and 36.6\%, keeps changes on English and the evaluated four-language European aggregate within 0.05\%, and retains 78.5\%/77.3\% of original model-vocabulary rows at the same IDs.
Constraint-matched global and frequency-based removal achieve similar Ukrainian compression but increase English/European token counts by 0.7--2.2\%; fresh same-size retraining compresses Ukrainian slightly more but retains effectively no same-ID rows and increases English token counts by 7.6--8.6\%.
The reallocation increases token counts on the evaluated three-language Cyrillic micro-aggregate by 6.7\%/10.1\%.
Structural audits find all 28,134/45,398 inserted BPE nodes reachable under ordinary rank-ordered merging and no retained same-ID model-vocabulary entry newly broken.
We release all tokenizers and code.\footnote{\url{https://github.com/BogdanDidenko/tokenizer-transfer-framework}; reproducibility release \texttt{tokshop-colm2026-repro-v1}.}
\end{abstract}

\section{Introduction}

Vocabulary adaptation for underrepresented languages and specialized domains differs fundamentally by tokenizer architecture.
For SentencePiece-based models, \citet{kiulian-etal-2025-english} demonstrate fixed-vocabulary reallocation for Ukrainian, Arabic, and Georgian via score-reassigned merging.
For byte-level BPE \citep{radford2019language}---used in GPT-OSS-20B \citep{openai-2025-gpt-oss}, Nemotron-3 \citep{nvidia2024nemotron}, Qwen \citep{yang-etal-2025-qwen3}, and Aya \citep{aryabumi-etal-2024-aya23}---to our knowledge no post-hoc fixed-vocabulary method yet solves this setting, despite high \textit{fertility} (tokens per word) in underrepresented languages: the unmodified Nemotron/GPT-OSS tokenizers require 2.71/2.60 tokens per Ukrainian word versus 1.32/1.26 for English (Table~\ref{tab:extended_metrics}), increasing inference cost and compressing effective context windows \citep{rust-etal-2021-good,petrov-etal-2024-language}.

Existing work on byte-level BPE expands the vocabulary, replaces the tokenizer, or changes allocation during pretraining; none addresses post-hoc adaptation under a fixed vocabulary budget (\S2).
We evaluate cheap same-size retraining directly; surgery instead preserves every non-selected vocabulary index and its associated input/output embedding rows.
This is a construction-time compatibility advantage, not a downstream-quality claim; measuring continued-pretraining savings requires a model-level experiment.
The structural obstacle is the \textbf{merge ordering problem}: in byte-level BPE, merge rules are applied greedily in rank order; when a new token is inserted with a heuristic split, an earlier-ranking merge may fire first, rendering the inserted token strict-unreachable---a failure mode termed \textit{ill-tokenization} \citep{balde-etal-2024-adaptive}.
\citet{balde-etal-2024-adaptive} report the same failure mode in domain vocabulary adaptation, and our worked Ukrainian example in \S\ref{sec:ordering} exhibits the corresponding target-rank conflict.
Our construction goal is therefore stricter than vocabulary membership: every inserted node must remain derivable by ordinary rank-ordered BPE, and no retained same-ID entry may lose that property.

We propose \textbf{tokenizer surgery}, centered on \textbf{BPE-guided insertion}.
We convert tokens from a \emph{character-level} BPE donor---whose merges operate on Unicode characters rather than the target's raw UTF-8 bytes---and register a decomposition reachable under the target merge ordering, without changing BPE inference.
To keep the vocabulary fixed \citep{kiulian-etal-2025-english,downey-etal-2023-embedding}, \textbf{script-aware removal} frees slots under a configurable retention filter; the target script is then rebuilt from a clean base inventory before donor-token insertion.
This instantiation is immediately useful when adaptation can be framed as writing-system-level reallocation, as in Ukrainian.
When the target language shares its script with languages that must also be preserved, the current pipeline lacks a reliable way to free slots without deleting from the same inventory it aims to keep; we leave such same-script reallocation to future work.

Our contributions:
\begin{enumerate}
\itemsep0em
    \item We formalize the merge ordering problem and propose target-guided insertion that makes every inserted node derivable under the target's merge ranks (\S\ref{sec:ordering}--\ref{sec:guided}).
    \item We combine dependency-safe script-aware removal, target-script base reconstruction, and target-guided insertion in a fixed-vocabulary writing-system-level pipeline (\S\ref{sec:method}).
    \item We evaluate real removal, continued-BPE, runtime-patching, and retraining baselines under a pinned suite, and audit inserted, retained, and full-vocabulary merge reachability (\S\ref{sec:experiments}).
\end{enumerate}

\section{Background and Related Work}

\paragraph{Vocabulary Adaptation.}
Approaches differ fundamentally in whether they expand, replace, or reallocate vocabulary.
\textit{Expansion} adds tokens to the existing vocabulary, increasing embedding matrix size: \citet{chau-etal-2020-parsing} extend multilingual BERT; \citet{kim-etal-2024-eeve} add 8,960 Korean tokens to SOLAR-10.7B with progressive unfreezing; \citet{wang-etal-2020-extending} extend to low-resource languages.
\textit{Full replacement} trains a new tokenizer and reinitializes all embeddings \citep{dobler-de-melo-2023-focus,remy-etal-2023-tiktotok,downey-etal-2023-embedding}.
\textit{Pre-training-time} methods modify BPE training itself: OBPE \citep{patil-etal-2022-overlap} biases merge selection toward cross-lingual overlap; XLM-V \citep{liang-etal-2023-xlm} trains a larger shared vocabulary.
Continued BPE training \citep{purason-etal-2025-teaching} extends an existing tokenizer by resuming merge learning on target-domain data and pairs this with leaf-based pruning, reducing unreachable or unused added tokens without relying on an auxiliary donor tokenizer; we use the authors' implementation directly.\footnote{\url{https://github.com/taidopurason/tokenizer-extension}, commit \texttt{15e9ed7f}.}
For SentencePiece tokenizers, \citet{kiulian-etal-2025-english} maintain fixed vocabulary size by merging score-reassigned vocabularies for Ukrainian, Arabic, and Georgian, but do not repair the explicit byte-level BPE merge tables studied here. SentencePiece provides both Unigram and BPE models; merge ancestry is recoverable for the latter, whereas score reassignment is native to the former.
In a same-script setting, \citet{ociepa-etal-2025-bielik} report that directly combining a Polish tokenizer with Mistral became ambiguous because overlapping vocabularies did not induce compatible token-pair merges.
Our setting is narrower and more structural: post-hoc modification of an existing byte-level BPE merge table under a fixed vocabulary budget.
\citet{land-bartolo-2024-fishing} independently document that many tokens in deployed multilingual tokenizers are under-trained, supporting the practical case for removal-based reallocation that targets rarely-trained inventory.
To our knowledge, no prior work addresses post-hoc token insertion in fixed-vocabulary byte-level BPE via an explicit structural correction of the stored merge table.

\paragraph{The Ill-Tokenization Problem.}
\citet{balde-etal-2024-adaptive} identify that naively appended domain tokens receive lower merge priority than existing vocabulary, causing BPE to ignore them---a failure they term \textit{ill-tokenization}.
Their solution, AdaptBPE, patches BPE initialization at runtime via longest-substring matching and does not modify the stored merge table; its published protocol expands rather than reallocates the vocabulary.
The study evaluates BART/RoBERTa (64.13\% ill-tokenization before the patch); a separate limitation diagnostic finds 27.76\% under standard LLaMA-2-7B tokenization but does not apply AdaptBPE to that model.
\citet{sharthak-etal-2025-tokenadapt} observe the same root cause and sidestep it via wholesale tokenizer transplantation, initializing the destination tokenizer's embedding matrix with local subword and global semantic heuristics.\footnote{\url{https://github.com/Tinycompany-AI/TokenAdapt}, commit \texttt{3f7ff1de}.} TokenAdapt is therefore an embedding-transfer method rather than an alternative tokenizer construction: its token counts are exactly those of its chosen destination tokenizer.
\citet{purason-etal-2025-teaching} likewise report that conventional tokenizer extension can add tokens that are unreachable or never used, and provide an open-source diagnostic for detecting unreachable vocabulary entries.
Our BPE-guided insertion addresses this failure structurally within the current writing-system-level pipeline by correcting the stored merge table rather than patching runtime behavior.

\paragraph{Ukrainian-language tokenization.}
Adjacent Ukrainian-NLP work targets different layers of the stack: rule-based word/sentence segmentation \citep{tokenize-uk-langek}, tokenizer-efficiency evaluation \citep{frontiers-2025-ukrainian-nlp}, and multilingual-LLM adaptation combining tokenizer reallocation with data curation and model training \citep{paniv-etal-2026-data}.
None addresses post-hoc fixed-vocabulary byte-level BPE merge-table modification, which is the contribution of this paper.

\paragraph{Embedding Initialization.}
Initializing embeddings for new tokens is orthogonal to our structural contribution but critical for downstream performance.
Methods range from sub-token averaging \citep{kim-etal-2024-eeve} and overlap-based transfer \citep{dobler-de-melo-2023-focus,remy-etal-2023-tiktotok,kiulian-etal-2025-english} to model-aware approaches such as MATT \citep{haltiuk-smywinski-pohl-2025-matt}.
Our tokenizer surgery is compatible with any initialization method applied post-surgery.

\section{Method}
\label{sec:method}

\subsection{Pipeline Overview}

Tokenizer surgery maintains a fixed vocabulary size $|V|$ in three stages: script-aware removal frees slots, target-script base reconstruction restores a controlled inventory of target-script \emph{base tokens}, and BPE-guided insertion adds donor tokens into the rebuilt byte-level vocabulary.
These stages address, respectively, collateral deletion outside the declared removal inventory, missing partial-byte dependencies after the target-script reset, and target-rank conflicts that can make simple donor-merge appends unreachable (\S\ref{sec:ordering}).

\subsection{Script-Aware Token Removal}

We classify each token by writing system using Unicode script detection via \texttt{unicodedataplus} and select removal candidates from a configurable set of removal-eligible scripts while excluding a declared \emph{retention inventory}. Tokens without a selected removal script, including Common-only surfaces, are not selected by this rule; the separate target-script reset takes precedence over the exclusion because it deliberately rebuilds the Cyrillic region. This differs from the corpus-frequency vocabulary reduction of \citet{alabi-etal-2022-adapting}: they retain high-frequency subwords from the adaptation corpora (plus 1,000 entries from the original tokenizer), whereas we classify decoded surfaces directly and validate the resulting merge structure.
In our tokenizer artifacts, higher vocabulary IDs generally correspond to later-added merges, so we use descending ID order as a practical removal heuristic within each removal script.
A decoded surface may contain characters from several writing systems, so it is excluded from script-selected removal if any character belongs to the retention inventory. This additional \texttt{unicodedataplus} check prevents a multi-script surface from being selected merely because one of its labels is removal-eligible.

In these experiments, quantitative collateral-preservation claims are limited to English and the four Latin-script corpora in the EU aggregate.
Additional build-specific exclusions in the frozen construction configs are reproducibility parameters, not evaluated preservation targets.
Appendix~\ref{app:replaced} reports every selected row by primary writing system; we make no preservation claim for writing systems absent from the evaluation suite.

After removal, we first apply \textbf{three-component checking}: a merge $[L, R] \to T$ is retained only if all three of $L$, $R$, and $T$ remain in the vocabulary; if any one was removed, the merge rule is dropped (no merge is added in this step).
This local check is necessary but not sufficient: removing an internal node can leave a retained parent without any complete derivation even when every remaining merge is individually well formed.
We therefore run an exact strict-backend merge-graph audit.
For an optional script-selected prerequisite that breaks a retained token, we restore that prerequisite and replace it with the next eligible descending-ID candidate; if a mandatory target-reset row caused the failure, the dependent token would enter the removal closure.
We repeat selection, merge cleanup, and strict audit until no retained same-ID entry is newly broken while preserving the exact slot budget.
This dependency repair restored 2 Nemotron and 14 GPT-OSS candidates; no target-reset closure was needed in either build.

\subsection{Target-Script Base Reconstruction}
\label{sec:base-rebuild}

For the target script itself---Cyrillic in our Ukrainian setting---we do not attempt to preserve arbitrary higher-level merges.
Byte-level BPE creates hidden dependencies through shared partial tokens (e.g., the byte token \texttt{[20 D0]}, a visible-space byte fused with a Cyrillic lead byte, participates in merges for many distinct Cyrillic letter forms), making selective retention brittle.
Instead, we reduce the target-script region to letter-level coverage and then complete a controlled base inventory of \emph{base tokens} containing every target-script letter in uppercase and lowercase, with and without leading space.
Existing \emph{base tokens} are retained when available; only missing ones are synthesized by adding the minimal byte-level merges needed to assemble them from byte-level tokens that already remain in the vocabulary.

For example, Figure~\ref{fig:base-rebuild} shows how existing byte-level tokens \texttt{[20 D0]} and \texttt{[B0]} can be merged into the space-prefixed \emph{base token} \texttt{[20 D0 B0]} ({\fontencoding{T2A}\selectfont \textvisiblespace а}).
Analogous merges complete the rest of the target-script inventory and provide the controlled foundation on which BPE-guided insertion builds larger tokens (\S\ref{sec:guided}).

\begin{center}
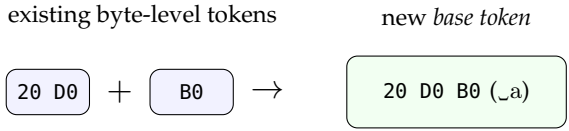

\begin{tikzpicture}[
    node distance=0.5cm and 0.65cm,
    frag/.style={draw, rounded corners, minimum width=1.1cm, minimum height=0.6cm, align=center, font=\small\ttfamily, fill=blue!5},
    tok/.style={draw, rounded corners, minimum width=2.9cm, minimum height=0.95cm, align=center, font=\small, fill=green!5},
    lab/.style={font=\footnotesize},
    arrow/.style={-{Stealth[length=2mm]}, thick}
]
\node[lab] at (0,1.0) {existing byte-level tokens};
\node[frag] (g1) at (-1.25,0) {\texttt{20 D0}};
\node[font=\large] at (-0.3,0) {$+$};
\node[frag] (g2) at (0.65,0) {\texttt{B0}};
\node[font=\large] at (1.65,0) {$\rightarrow$};
\node[tok] (letter) at (4.15,0) {\texttt{20 D0 B0}\hspace{0.35em}({\fontencoding{T2A}\selectfont \textvisiblespace а})};
\node[lab] at (4.15,1.0) {new \emph{base token}};
\end{tikzpicture}
\captionof{figure}{Composing a missing target-script \emph{base token} from existing byte-level tokens. If the inputs for a missing \emph{base token} already remain in the vocabulary, we add only the final merge that composes it; repeating this restores the target-script building blocks used for later insertion.}
\label{fig:base-rebuild}
\end{center}

\subsection{The Merge Ordering Problem}
\label{sec:ordering}

After reconstructing a controlled target-script inventory, the remaining question is no longer whether the pieces of a donor token exist.
It is whether the current target merge table will ever expose the boundary at which a newly added merge is supposed to fire.

\paragraph{Hidden internal nodes.}
This is more subtle than ordinary string segmentation because byte-level BPE already contains internal graph nodes that do not align with Unicode character boundaries.
In GPT-OSS, for example, \texttt{[20 D0]} (visible space plus a dangling Cyrillic lead byte) is the parent of space-prefixed letters such as \texttt{[20 D0 B0]} ({\fontencoding{T2A}\selectfont \textvisiblespace а}), \texttt{[20 D0 B1]} ({\fontencoding{T2A}\selectfont \textvisiblespace б}), and \texttt{[20 D0 BC]} ({\fontencoding{T2A}\selectfont \textvisiblespace м}).
The vocabulary also contains \texttt{[D0 BE D0]}, which combines the full letter {\fontencoding{T2A}\selectfont о} with the lead byte of the next Cyrillic character and is then extended into \texttt{[D0 BE D0 B2]} ({\fontencoding{T2A}\selectfont ов}), \texttt{[D0 BE D0 B4]} ({\fontencoding{T2A}\selectfont од}), \texttt{[D0 BE D0 BD]} ({\fontencoding{T2A}\selectfont он}), and related nodes.

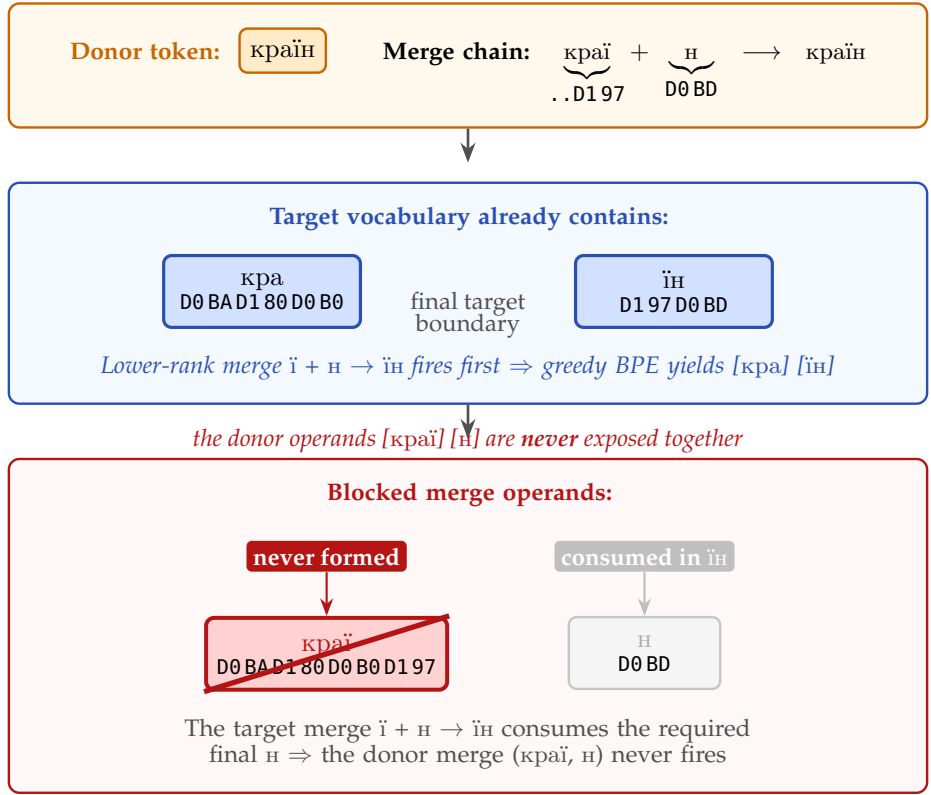
\begin{figure*}[t]
\centering
\definecolor{donorcolor}{RGB}{200,100,0}
\definecolor{tgtcolor}{RGB}{40,80,180}
\definecolor{badcolor}{RGB}{180,20,20}
\definecolor{textgray}{RGB}{80,80,80}
\definecolor{lightorange}{RGB}{255,235,200}
\definecolor{lightblue}{RGB}{210,225,255}
\definecolor{lightred}{RGB}{255,210,210}
\begin{tikzpicture}[
  font=\small,
  tgtbox/.style={
    rectangle, rounded corners=3pt,
    fill=lightblue, draw=tgtcolor, line width=1.2pt,
    minimum width=2.6cm, minimum height=0.95cm,
    inner sep=4pt, align=center
  },
  badbox/.style={
    rectangle, rounded corners=3pt,
    fill=lightred, draw=badcolor, line width=1.5pt,
    minimum width=2.6cm, minimum height=0.95cm,
    inner sep=4pt, align=center
  },
  flowArrow/.style={-Stealth, line width=1.0pt, draw=textgray},
  smBadge/.style={rectangle, rounded corners=2pt, inner sep=2.5pt, font=\footnotesize\bfseries},
  secbox/.style={rectangle, rounded corners=4pt, inner sep=9pt, align=center, line width=0.9pt, text width=11.5cm},
  node distance=0.65cm
]
\node[secbox, fill=lightorange!35, draw=donorcolor] (box1) at (0,0)
  {{\bfseries\textcolor{donorcolor}{Donor token:}}\;
   \tikz[baseline=-3pt]{\node[rectangle, rounded corners=3pt, fill=lightorange, draw=donorcolor, line width=0.9pt, inner sep=4pt, font=\normalsize\bfseries]{\cy{країн}};}
   \qquad
   {\bfseries Merge chain:}\;
   $\underbrace{\text{\cy{краї}}}_{\text{\footnotesize\texttt{..D1\,97}}}
   +\;
   \underbrace{\text{\cy{н}}}_{\text{\footnotesize\texttt{D0\,BD}}}
   \;\longrightarrow\;
   \text{\cy{країн}}$};

\draw[flowArrow] (box1.south) -- ++(0,-0.45cm);

\node[secbox, fill=lightblue!25, draw=tgtcolor, below=0.7cm of box1] (box2)
  {{\bfseries\textcolor{tgtcolor}{Target vocabulary already contains:}}\\[8pt]
   \begin{tikzpicture}[baseline=(current bounding box.center), font=\small, inner sep=0pt]
     \node[tgtbox] (kra) {\normalsize\bfseries\cy{кра}\\[-1pt]
       {\footnotesize\texttt{D0\,BA\,D1\,80\,D0\,B0}}};
     \node[tgtbox, right=2.8cm of kra] (yin) {\normalsize\bfseries\cy{їн}\\[-1pt]
       {\footnotesize\texttt{D1\,97\,D0\,BD}}};
     \node[font=\footnotesize, color=textgray, align=center] at ($(kra.east)!0.5!(yin.west)$) {final target\\[-2pt]boundary};
   \end{tikzpicture}\\[6pt]
   \textcolor{tgtcolor}{\small\itshape Lower-rank merge \cy{ї} + \cy{н} $\to$ \cy{їн} fires first $\Rightarrow$ greedy BPE yields [\cy{кра}] [\cy{їн}]}};

\draw[flowArrow] (box2.south) -- ++(0,-0.45cm);
\node[below=0.47cm of box2, anchor=center, font=\small\itshape, color=badcolor]
  {the donor operands [\cy{краї}] [\cy{н}] are \textbf{never} exposed together};

\node[secbox, fill=lightred!15, draw=badcolor, below=0.7cm of box2] (box3)
  {{\bfseries\textcolor{badcolor}{Blocked merge operands:}}\\[12pt]
   \begin{tikzpicture}[baseline=(kraibox.center), font=\small, inner sep=0pt]
     \node[smBadge, fill=badcolor, text=white] (nvr) at (0, 1.0cm) {never formed};
     \node[badbox] (kraibox) at (0, 0)
       {\normalsize\bfseries\textcolor{badcolor}{\cy{краї}}\\[-1pt]
        {\footnotesize\texttt{D0\,BA\,D1\,80\,D0\,B0\,D1\,97}}};
     \draw[-Stealth, badcolor, line width=0.8pt] (nvr.south) -- (kraibox.north);
     \draw[badcolor, line width=2pt] (kraibox.south west) -- (kraibox.north east);
     \node[smBadge, fill=gray!50, text=white] (unr) at (4.2cm, 1.0cm) {consumed in \cy{їн}};
     \node[rectangle, rounded corners=3pt, fill=gray!8, draw=gray!45, line width=0.9pt,
       minimum width=2.0cm, minimum height=0.95cm, inner sep=4pt, align=center] (nbox) at (4.2cm, 0)
       {\normalsize\bfseries\textcolor{gray!65}{\cy{н}}\\[-1pt]
        {\footnotesize\texttt{D0\,BD}}};
     \draw[-Stealth, gray!50, line width=0.8pt] (unr.south) -- (nbox.north);
   \end{tikzpicture}\\[8pt]
   \textcolor{textgray}{\small The target merge \cy{ї} + \cy{н} $\to$ \cy{їн} consumes the required final \cy{н} $\Rightarrow$ the donor merge (\cy{краї}, \cy{н}) never fires}};
\end{tikzpicture}
\caption{The \textbf{merge ordering problem} for the token \cy{країн}. The donor tokenizer contains the merge (\cy{краї}, \cy{н}) $\to$ \cy{країн}, but both targets apply the lower-rank merge \cy{ї} + \cy{н} $\to$ \cy{їн} first and expose [\cy{кра}] [\cy{їн}]. The required donor operands [\cy{краї}] [\cy{н}] therefore never coexist, so directly copying the donor merge cannot produce the inserted token.}
\label{fig:merge-ordering}
\end{figure*}

\paragraph{Worked donor-target conflict.}
A merge $(L, R) \to T$ is therefore reachable only if tokenizing $T$ under the current merge ranks naturally yields $[L, R]$ as the final unresolved split for $T$.
Figure~\ref{fig:merge-ordering} gives a worked example for the Ukrainian token {\fontencoding{T2A}\selectfont країн}.
In both GPT-OSS and Nemotron, the unmodified target tokenizer already routes this string through \texttt{[D0 BA D1 80 D0 B0] $|$ [D1 97 D0 BD]}, i.e., roughly {\fontencoding{T2A}\selectfont кра}$\mid${\fontencoding{T2A}\selectfont їн}.
The second node \texttt{[D1 97 D0 BD]} is itself built as \texttt{[D1 97] + [D0 BD]} and reused in larger merges such as {\fontencoding{T2A}\selectfont Україн}.
A donor-consistent split would place the boundary after {\fontencoding{T2A}\selectfont краї}, but if we register that donor boundary, the new merge never fires because the current target graph does not expose \texttt{[D0 BA D1 80 D0 B0 D1 97]} as the final unresolved left component.
The added token is present in the vocabulary but strict-unreachable through the merge graph.

\paragraph{Definition and validation protocol.}
We call this the \textbf{merge ordering problem}: a token can be present in the vocabulary yet remain strict-unreachable because the merge registered for it disagrees with the split induced by the current rank-ordered merge table.
\textit{Strict merge reachability} asks whether ordinary rank-ordered BPE reduces a serialized model-vocabulary key to exactly its own ID.
The audit calls the BPE model directly, bypassing the pre-tokenizer and added-token matcher, so it tests the stored merge graph rather than vocabulary membership alone.
Unless explicitly labeled operational in the auxiliary runtime analysis (\S\ref{sec:runtime-lookup}), ``reachable'' below means strict merge reachability; every reported fraction uses checked entries as its denominator and lists decoding exclusions separately.

\subsection{BPE-Guided Insertion}
\label{sec:guided}

After target-script base reconstruction (\S\ref{sec:base-rebuild}), the rebuilt tokenizer already contains a controlled byte-level inventory of target-script \emph{base tokens}.
For each donor surface $T$, we convert it to the target's byte-level representation and simulate the \emph{current target} merge table.
If the reachable decomposition contains more than two pieces, we recursively materialize an adjacent intermediate reachable under those same target ranks, re-simulate $T$, and repeat.
Once the current decomposition is $(A,B)$, we register $(A,B)\to T$ at the end of the target merge table.
Thus every inserted merge agrees with the split that greedy target-side BPE actually exposes; the donor merge hierarchy is neither copied nor required.

The insertion of one donor surface is transactional.
All target-side prerequisite nodes required to make $T$ reachable consume replacement slots; if the full chain does not fit, its tentative nodes are rolled back and the next donor candidate is tried.
Insertion stops only when the budget freed by script-aware removal, target-script reset, and base reconstruction is exhausted.

\begin{algorithm}[t]
\caption{Budgeted Target-Guided Donor Insertion}
\label{alg:insert}
\small
\begin{algorithmic}
\REQUIRE ordered donor surfaces $\Omega$, target merge ranks $R$, slot budget $B$
\STATE $\mathit{added} \gets 0$
\FORALL{$T \in \Omega$}
    \IF{$\mathit{added} = B$}
        \STATE \textbf{stop}
    \ENDIF
    \STATE $C \gets \textsc{TargetClosure}(T,R)$ \COMMENT{each merge follows the split reachable under $R$}
    \IF{$\mathit{added} + |C| \leq B$}
        \STATE \textsc{InsertByteMerges}$(C)$ and update $R$
        \STATE $\mathit{added} \gets \mathit{added} + |C|$
    \ENDIF
\ENDFOR
\STATE \textbf{return} inserted target-reachable closure under budget $B$
\end{algorithmic}
\end{algorithm}

\paragraph{Donor requirements.}
For correctness, the donor need only expose decodable Unicode token surfaces; it need not expose its merge graph or share vocabulary IDs, special-token assignments, vocabulary size, or byte encoding with the target.
For useful adaptation, it should be trained on the target language or domain and supply enough candidates to use the available slot budget.
Candidate order is an explicit input to Algorithm~\ref{alg:insert}, not a correctness condition: our Ukrainian experiments use ascending serialized donor-vocabulary ID only as a deterministic order and do not interpret it as corpus frequency.

Unlike AdaptBPE's runtime patch for appended domain vocabulary \citep{balde-etal-2024-adaptive}, our approach is a structural fix for writing-system-level tokenizer replacement: we correct the merge table itself, so standard BPE implementations require no modification.

\section{Experiments}
\label{sec:experiments}

\subsection{Setup}

We evaluate two Ukrainian tokenizer adaptations based on \textbf{Nemotron-3-Nano} \citep{nvidia2024nemotron} (131,072 BPE model-vocabulary entries) and \textbf{GPT-OSS-20B} \citep{openai-2025-gpt-oss} (199,998 BPE model-vocabulary entries).
Donor tokens come from the Lapa project's Unicode-character BPE Ukrainian tokenizer release \citep{lapa-tokenizer-release-2025}, whose \texttt{uk\_tokenizer.json} contains 150,000 entries.
For these Ukrainian builds, donor eligibility retains stripped surfaces of at least two Unicode code points whose non-space characters are all Cyrillic; leading-space and non-leading-space forms remain distinct.
We additionally require the surface to remain one piece under the target tokenizer's pre-tokenizer, then traverse the eligible list in ascending donor-vocabulary ID.
These are script and boundary filters over a Ukrainian-trained donor, not a language classifier or a corpus-frequency ranking.
The script-selection configs and resulting replacement manifests are pinned; Appendix~\ref{app:replaced} gives their complete primary-writing-system breakdown.
We interpret collateral preservation only on the evaluated corpora rather than inferring it from whether a writing system was excluded from row selection.
We report tokenizer-level fertility (tokens per whitespace-delimited word) on a pinned suite: 100,000 documents per subset for Ukrainian Malyuk, C4 English, a four-language EU aggregate, and a three-language Cyrillic aggregate, plus the 94 available Cyrillic Crimean-Tatar documents. Appendix~\ref{app:extended_metrics} reports the exact dataset codes and per-subset counts.
The samples, dataset revisions, shuffle order (100,000-example buffer, seed 42), text-sequence hashes, and evaluation runtime (Transformers 4.57.3; Tokenizers 0.22.1) are pinned.
EU is the micro-aggregate of Spanish, French, Italian, and German; the broader evaluation sheet is given in Appendix~\ref{app:extended_metrics}.

\subsection{Baselines}
\label{sec:baselines}

All fixed-budget surgery arms use the same base tokenizer, donor order, target pre-tokenizer filter, and replacement budgets (28,134 Nemotron rows; 45,398 GPT-OSS rows).
The script-aware, global-ID-after-reset, and Alabi-style selector arms also use the same strict dependency audit; optional prerequisite rows are restored and replaced under each arm's own ranking.
The Alabi-style selector, Purason, and the direct-transfer control use the same pinned 100M-character Ukrainian HPLT slice.
The AdaptBPE arms instead draw eligible surfaces in the same order from the released donor used by our method.
The fresh-retraining baseline instead uses a separately pinned 100M-character UK50/EU50 mixture drawn from HPLT Ukrainian and five Wikipedia editions; none of its source datasets is used in the evaluation suite.
We compare:
(i) global descending-ID removal after the same target reset;
(ii) an Alabi-style selector \citep{alabi-etal-2022-adapting} that replaces the low-frequency complement of adaptation-corpus tokens while protecting the first 1,000 original IDs except where the shared target-script reset must take precedence;
(iii) continued BPE with leaf-frequency pruning from the official Purason et al.\ implementation \citep{purason-etal-2025-teaching};
(iv) fresh same-size BPE retraining on the declared UK50/EU50 mixture; and
(v) AdaptBPE \citep{balde-etal-2024-adaptive}.
For AdaptBPE, whose released code modifies only slow RoBERTa/BART tokenizers,\footnote{\url{https://github.com/gb-kgp/adaptbpe}, commit \texttt{949556e4}.} we port the published recursive longest-substring initialization to each target's pinned pre-tokenizer and merge table. We replace its Latin-only single-character helper set with the existing 256-symbol byte alphabet so Ukrainian byte strings admit a complete cover, and verify that the optimized port matches a direct transcription of the published recursion on 500 Ukrainian pre-tokenized pieces per target (0 mismatches).
We report both a constraint-matched operationalization that reuses precisely the script-aware replacement IDs without adding merge rules and the published expansion setting, which adds the same number of donor types and is explicitly marked as outside the fixed-vocabulary constraint.
The fixed runtime variant can assign one donor surface to every freed row because it pays no prerequisite-node cost, but it requires a non-standard tokenizer at inference; our structural method consumes some rows on prerequisites and runs in an unmodified BPE implementation.

We additionally run a \textbf{direct target-native merge-transfer control} for reachability rather than fertility.
For each target, we train a fresh byte-level BPE with that target's own pre-tokenizer on the same 100M-character Ukrainian slice, reuse exactly the script-aware arm's replacement IDs, clean invalidated target merges, and append an ancestry-closed donor merge prefix in donor-rank order without target-guided repair.
Because this control changes the donor inventory and candidate ranking, its fertility gap from our released-donor build is not a causal estimate of BPE-guided insertion; its purpose is the auxiliary runtime analysis in \S\ref{sec:runtime-lookup}.

TokenAdapt \citep{sharthak-etal-2025-tokenadapt} is not listed as a separate token-count row because it initializes model embeddings for an already chosen destination tokenizer; it does not change that tokenizer's segmentation.
A valid TokenAdapt comparison is model-level (initialization, equal continued-pretraining compute, and downstream/perplexity evaluation), for which the complete 20B/30B weights are outside this tokenizer-only experiment.

\subsection{Language Adaptation Results}

\begin{table}[t]
\centering
\small
\resizebox{\columnwidth}{!}{%
\begin{tabular}{@{}lccccc@{}}
\toprule
\textbf{Tokenizer} & \textbf{UK} & \textbf{EN} & \textbf{EU} & \textbf{Donor surfaces} & \textbf{IDs kept} \\
\midrule
Nemotron & 2.71 & 1.32 & 1.84 & --- & 100\% \\
\textbf{Script-aware (ours)} & \textbf{1.80} & 1.32 & 1.84 & 27,975 & 78.5\% \\
\midrule
GPT-OSS & 2.60 & 1.26 & 1.69 & --- & 100\% \\
\textbf{Script-aware (ours)} & \textbf{1.65} & 1.26 & 1.69 & 45,232 & 77.3\% \\
\bottomrule
\end{tabular}
}
\caption{Fertility (tokens/whitespace-delimited word, $\downarrow$). ``Donor surfaces'' counts inserted donor surfaces, not vocabulary expansion; prerequisite BPE nodes also consume replacement slots. ``IDs kept'' is the fraction of original model-vocabulary rows retaining the same token at the same numeric ID.}
\label{tab:results}
\end{table}

\begin{table*}[t]
\centering
\scriptsize
\setlength{\tabcolsep}{4pt}
\begin{tabular}{@{}llcrrrrr@{}}
\toprule
\textbf{Target} & \textbf{Method} & \textbf{Budget} & \textbf{IDs kept} & $\Delta$\textbf{UK} & $\Delta$\textbf{EN} & $\Delta$\textbf{EU} & $\Delta$\textbf{Cyr. agg.} \\
\midrule
Nemotron & Ours: script-aware & fixed & 78.5\% & -33.48\% & +0.03\% & +0.01\% & +6.65\% \\
 & Global last-ID after reset & fixed & 78.5\% & -33.41\% & +0.97\% & +2.08\% & +6.74\% \\
 & Alabi-style frequency & fixed & 78.5\% & -33.44\% & +0.82\% & +1.96\% & +6.67\% \\
 & AdaptBPE runtime & fixed & 78.5\% & -27.20\% & +0.07\% & +0.02\% & +33.16\% \\
 & Purason leaf-frequency & fixed & 74.3\% & -32.78\% & +1.18\% & +2.81\% & -5.38\% \\
 & Same-size BPE (UK50/EU50) & fixed & 0.0\% & -36.42\% & +8.61\% & -0.44\% & +3.23\% \\
 & AdaptBPE runtime & +28,134 & 100.0\% & -32.89\% & +0.00\% & +0.00\% & +2.45\% \\
\midrule
GPT-OSS & Ours: script-aware & fixed & 77.3\% & -36.60\% & +0.05\% & +0.02\% & +10.06\% \\
 & Global last-ID after reset & fixed & 77.3\% & -36.56\% & +0.78\% & +2.20\% & +10.10\% \\
 & Alabi-style frequency & fixed & 77.3\% & -36.58\% & +0.70\% & +2.12\% & +10.06\% \\
 & AdaptBPE runtime & fixed & 77.3\% & -30.08\% & +0.09\% & +0.03\% & +39.47\% \\
 & Purason leaf-frequency & fixed & 73.9\% & -35.71\% & +1.06\% & +3.17\% & -4.10\% \\
 & Same-size BPE (UK50/EU50) & fixed & 0.0\% & -38.16\% & +7.64\% & -0.77\% & +7.14\% \\
 & AdaptBPE runtime & +45,398 & 100.0\% & -36.55\% & +0.00\% & +0.00\% & +2.76\% \\
\bottomrule
\end{tabular}
\caption{Token-count changes relative to each unmodified base on the pinned evaluation suite (lower is better). ``fixed'' means the same total active tokenizer-ID budget as the corresponding base (BPE model vocabulary plus any added-token layer), not preservation of the old assignments. The first three rows in each block share the target-script reset, donor order, insertion procedure, and slot budget; only the selector for non-Cyrillic fill slots differs. EU is the ES/FR/IT/DE micro-aggregate; Cyr. agg. is the three-language Cyrillic micro-aggregate. ``IDs kept'' is the share of original model-vocabulary rows retaining the same token at the same numeric ID; coincidental matches after fresh retraining round to 0.0\%. The final AdaptBPE row in each block is its published expansion setting and is not fixed-vocabulary.}
\label{tab:baselines}
\end{table*}

Table~\ref{tab:results} shows Ukrainian fertility falling from 2.71 to 1.80 and from 2.60 to 1.65, while English/EU fertility is unchanged at the displayed precision. The row audit preserves every non-selected entry: 102,938/131,072 and 154,600/199,998 original token-to-ID assignments remain intact.

The cross-language Cyrillic cost is material: the three-language micro-aggregate changes by +6.65\%/+10.06\%, with per-subset results reported in Appendix~\ref{app:extended_metrics}. Because these subsets share Cyrillic with the target inventory, they are affected by the mandatory target reset; reallocation cannot be uniformly Pareto-optimal across inventories that share the same writing system.

Table~\ref{tab:baselines} gives the baseline comparisons.
At fixed target reset, global descending-ID removal matches our Ukrainian and Cyrillic-aggregate changes within 0.10 points but adds +0.97\%/+2.08\% EN/EU tokens on Nemotron and +0.78\%/+2.20\% on GPT-OSS.
Its retained-entry audit passes, identifying script filtering as a preservation policy rather than a reachability requirement.
Alabi-style selection also matches Ukrainian but costs +0.82\%/+1.96\% and +0.70\%/+2.12\% EN/EU, versus at most +0.05\% for ours.

The fixed-budget AdaptBPE arm combines the shared target reset, which removes base Cyrillic entries and incident merges, with runtime substring initialization. It changes Ukrainian by $-27.20\%$/$-30.08\%$ and the Cyrillic aggregate by +33.16\%/+39.47\%.
The published expansion arm retains the base inventory and reaches $-32.89\%$/$-36.55\%$ Ukrainian, while the aggregate changes by +2.45\%/+2.76\%. Its recursive initialization can select a new donor token inside a larger pre-tokenized piece; because the original merge table has no rules using that new atom, a base token spanning it may no longer form. Expansion adds 28,134/45,398 entries and requires modified inference.

Purason continued BPE reaches $-32.78\%$/$-35.71\%$ Ukrainian but costs +1.18\%/+1.06\% English and +2.81\%/+3.17\% EU, retaining 74.3\%/73.9\% of IDs.
Fresh UK50/EU50 retraining uses the same total active tokenizer-ID budget as each base but does not preserve its assignments; it compresses Ukrainian still more ($-36.42\%$/$-38.16\%$), raises English by 8.61\%/7.64\%, and retains effectively 0\% of original model-vocabulary rows at the same IDs.
Surgery's measured advantage is thus index compatibility and selected-language segmentation, not cheap training or maximal Ukrainian compression.

\paragraph{Verification.}
Direct Rust-model audits find 0/28,134 and 0/45,398 broken inserted nodes.
When the full pipeline is forced through the ordinary merge path, it emits all 27,975/45,232 donor surfaces at their assigned IDs with exact roundtrips.
Strict audits find 0/101,938 and 0/154,600 newly broken retained same-ID entries and 0/130,072 and 0/199,998 broken full model-vocabulary entries (excluding 1,000 Nemotron added-token contents, checked operationally).
The global and Alabi-style arms likewise introduce no retained failures; Appendix~\ref{app:stress-tests} separates vocabulary membership, strict model reachability, and operational surface reachability.

\subsection{Auxiliary Runtime Analysis: Exact Whole-Piece Lookup}
\label{sec:runtime-lookup}

The target implementation exposes a pre-merge shortcut separate from the BPE graph.
In Hugging Face Tokenizers BPE, the \texttt{ignore\_merges} option controls this \textit{exact whole-piece vocabulary lookup}: if an entire pre-tokenized piece is already a vocabulary entry, the model returns its ID before applying BPE merges; otherwise it proceeds with ordinary merging.\footnote{\url{https://huggingface.co/docs/tokenizers/main/api/models}}
Both evaluated target configurations enable this option.
For this diagnostic, \textit{operational isolated reachability} asks whether the full tokenizer pipeline encodes an isolated raw surface as exactly its assigned ID under a declared lookup state.

The direct target-native merge-transfer control leaves 4,513/28,134 Nemotron nodes (16.04\%) and 104/45,398 GPT-OSS nodes (0.23\%) strict-broken.
Among valid UTF-8 isolated surfaces, disabling the shortcut exposes 4,512/28,126 and 104/45,387 exact-ID failures; enabling it recovers every one, reducing isolated operational failures to 0/28,126 and 0/45,387.
Thus the shortcut masks 100\% of the observed isolated-surface failures in this control without repairing the merge graph.
It matches a complete pre-tokenized piece, not arbitrary occurrences inside a larger piece, so this runtime diagnostic complements rather than replaces structural validation.
Appendix~\ref{app:stress-tests} reports denominators and decoding exclusions.

\section{Conclusion}

Taken together, the results position tokenizer surgery as a compatibility-preserving alternative to full replacement rather than a route to maximal target-language compression.
Its central guarantee is that inserted nodes remain derivable under ordinary BPE while retained same-ID entries remain structurally valid.
The auxiliary whole-piece control shows why default isolated-token behavior must be reported separately from that guarantee.
The measured cross-language Cyrillic degradation also shows that near-zero changes on the evaluated Latin-script suite do not imply blanket language preservation.

\section*{Ethics Statement}

The Ukrainian experiments use writing-system-level slot reallocation under a fixed vocabulary budget.
Only the English/EU Latin-script suite is evaluated as collateral preservation; the complete row-selection breakdown is reported separately as a construction record in Appendix~\ref{app:replaced}.
Whether an unevaluated writing system is excluded from row selection is an implementation parameter, not evidence of language-level preservation or a normative ranking of languages and cultures.
The observed degradation on the evaluated Cyrillic suite makes the shared-capacity trade-off explicit.

The donor-side data mixture also includes Cyrillic Crimean Tatar material.
We view support for the languages of Indigenous peoples of Ukraine and other communities historically connected with Ukraine as an important direction for future tokenizer and model development.
More broadly, we regard writing-system-aware adaptation as a tool for improving representation of under-served language groups under fixed-vocabulary constraints; such methods should be used to expand language support, not to justify the marginalization or erasure of particular languages, scripts, or speech communities.

\section*{Limitations}

\noindent\textbf{Compression vs. downstream performance.}
Our primary metric is tokenizer fertility (compression), which is an efficiency signal rather than a downstream quality guarantee.
Recent work shows that better compression does not uniformly predict task performance: \citet{lotz-etal-2025-beyond} find a statistically significant \textit{negative} correlation ($\rho{=}{-}0.59$) between compression and multiple-choice benchmark scores, and \citet{wegmann-etal-2025-tokenization} show the correlation can invert by task type.
We therefore treat lower fertility as evidence of tokenizer-level efficiency---reduced context-window consumption and faster generation---without claiming uniform downstream gains.
Structural \emph{reachability} is a separate construction-time property: a tokenizer can compress well while containing vocabulary entries that its merge graph cannot derive, even if the default runtime emits some exact whole pieces.
Our builds satisfy strict reachability for all inserted and retained model entries and operational reachability for all donor surfaces through ordinary rank-ordered merging; this invariant does not itself imply downstream gains.

\noindent\textbf{Runtime lookup scope.}
Both evaluated targets enable the whole-piece lookup defined in \S\ref{sec:runtime-lookup}, which recovers every valid isolated strict-broken surface in the direct-transfer control.
Consequently, our structural result should not be interpreted as evidence that direct transfer has the same operational failure rate under the targets' default settings, nor that every repaired node improves corpus fertility.
The guarantee concerns behavior without this lookup and makes the serialization robust to runtime settings; quantifying benefits for occurrences inside larger pre-tokenized pieces requires a context-distribution analysis with matched donor inventories and candidate order.

\noindent\textbf{Writing-system scope.}
The current fixed-vocabulary pipeline is strongest when adaptation can be framed as writing-system-level reallocation.
In same-script settings, such as adapting one Latin-script language while preserving another, the current slot-acquisition strategy would have to delete from the same inventory it seeks to preserve.
Our Alabi-style baseline applies corpus-frequency filtering \citep{alabi-etal-2022-adapting} under the same replacement budget, target-script reset, donor order, and insertion procedure. This tests one such selector, but same-script adaptation remains harder because corpus frequency need not preserve every language or domain that shares the script.

\noindent\textbf{Preservation scope.}
Our preservation claims are limited to tokenizer-level behavior on the evaluated corpora.
We make no behavioral claim for unevaluated writing systems, irrespective of whether the frozen construction config excludes their rows from removal.
The broader appendix tables show that Ukrainian-centric reallocation is not neutral for all nearby Cyrillic settings, so the current pipeline should not be read as uniformly preserving all neighboring languages.

\noindent\textbf{Segmentation-distribution shift.}
The EN/EU fertility increases from non-script-aware removal measure only additional fragmentation, not its full possible effect on the pretrained model.
Whenever the original tokenizer emitted a removed surface as one token, pretraining did not expose that occurrence under the new multi-token decomposition.
The component tokens may be individually familiar from other contexts, but their composition is not guaranteed to recover the removed token's learned representation.
We therefore hypothesize that indiscriminate removal can cause model-level degradation beyond the measured token-count increase; establishing its magnitude requires collateral-language perplexity and downstream benchmarks under a controlled embedding-initialization and continued-pretraining protocol.

\noindent\textbf{Embedding initialization.}
Our method modifies only the tokenizer; useful model behavior still depends on embedding initialization and continued pre-training.
TokenAdapt \citep{sharthak-etal-2025-tokenadapt} and MATT \citep{haltiuk-smywinski-pohl-2025-matt} are applicable at this subsequent model-level stage. They cannot be distinguished by the tokenizer-only counts reported here: TokenAdapt leaves the selected destination tokenizer unchanged, while changing how its embedding rows are initialized.

\FloatBarrier
\bibliographystyle{colm2026_conference}
\bibliography{custom}

\begin{thebibliography}{29}
\providecommand{\natexlab}[1]{#1}
\providecommand{\url}[1]{\texttt{#1}}
\expandafter\ifx\csname urlstyle\endcsname\relax
  \providecommand{\doi}[1]{doi: #1}\else
  \providecommand{\doi}{doi: \begingroup \urlstyle{rm}\Url}\fi

\bibitem[Alabi et~al.(2022)Alabi, Adelani, Mosbach, and
  Klakow]{alabi-etal-2022-adapting}
Jesujoba~O. Alabi, David~Ifeoluwa Adelani, Marius Mosbach, and Dietrich Klakow.
\newblock Adapting pre-trained language models to {A}frican languages via
  multilingual adaptive fine-tuning.
\newblock In \emph{Proceedings of the 29th International Conference on
  Computational Linguistics}, pp.\  4336--4349, Gyeongju, Republic of Korea,
  2022. International Committee on Computational Linguistics.
\newblock URL \url{https://aclanthology.org/2022.coling-1.382/}.

\bibitem[Aryabumi et~al.(2024)Aryabumi, Dang, Talupuru, Dash, Cairuz,
  et~al.]{aryabumi-etal-2024-aya23}
Viraat Aryabumi, John Dang, Dwarak Talupuru, Saurabh Dash, Hangyu Cairuz,
  et~al.
\newblock Aya expanse: Combining research breakthroughs for a new multilingual
  frontier.
\newblock \emph{arXiv preprint}, 2024.

\bibitem[Balde et~al.(2024)Balde, Roy, Mondal, and
  Ganguly]{balde-etal-2024-adaptive}
Gunjan Balde, Soumyadeep Roy, Mainack Mondal, and Niloy Ganguly.
\newblock Adaptive {BPE} tokenization for enhanced vocabulary adaptation in
  finetuning pretrained language models.
\newblock In \emph{Findings of EMNLP}, pp.\  14724--14733, 2024.
\newblock \doi{10.18653/v1/2024.findings-emnlp.863}.
\newblock URL \url{https://aclanthology.org/2024.findings-emnlp.863/}.

\bibitem[Chau et~al.(2020)Chau, Lin, and Smith]{chau-etal-2020-parsing}
Ethan~C. Chau, Lucy~H. Lin, and Noah~A. Smith.
\newblock Parsing with multilingual {BERT}, a small corpus, and a small
  treebank.
\newblock In \emph{Findings of EMNLP}, pp.\  1324--1334, 2020.
\newblock \doi{10.18653/v1/2020.findings-emnlp.118}.

\bibitem[Dobler \& de~Melo(2023)Dobler and de~Melo]{dobler-de-melo-2023-focus}
Konstantin Dobler and Gerard de~Melo.
\newblock {FOCUS}: Effective embedding initialization for monolingual
  specialization of multilingual models.
\newblock In \emph{Proceedings of EMNLP}, 2023.

\bibitem[Downey et~al.(2023)Downey, Blevins, Goldfine, and
  Steinert-Threlkeld]{downey-etal-2023-embedding}
C.~M. Downey, Terra Blevins, Nora Goldfine, and Shane Steinert-Threlkeld.
\newblock Embedding structure matters: Comparing methods to adapt multilingual
  vocabularies to new languages.
\newblock In \emph{Proceedings of the 3rd Workshop on Multi-lingual
  Representation Learning}, pp.\  268--281, 2023.

\bibitem[Haltiuk \& Smywinski-Pohl(2025)Haltiuk and
  Smywinski-Pohl]{haltiuk-smywinski-pohl-2025-matt}
Mykola Haltiuk and Aleksander Smywinski-Pohl.
\newblock Model-aware tokenizer transfer.
\newblock \emph{arXiv preprint arXiv:2510.21954}, 2025.

\bibitem[Kim et~al.(2024)Kim, Choi, and Jeong]{kim-etal-2024-eeve}
Seungduk Kim, Seungtaek Choi, and Myeongho Jeong.
\newblock Efficient and effective vocabulary expansion towards multilingual
  large language models.
\newblock \emph{arXiv preprint arXiv:2402.14714}, 2024.

\bibitem[Kiulian et~al.(2025)Kiulian, Polishko, Khandoga, Kostiuk,
  et~al.]{kiulian-etal-2025-english}
Artur Kiulian, Anton Polishko, Mykola Khandoga, Yevhen Kostiuk, et~al.
\newblock From {E}nglish-centric to effective bilingual: {LLM}s with custom
  tokenizers for underrepresented languages.
\newblock In \emph{Proceedings of the Fourth Ukrainian Natural Language
  Processing Workshop}, pp.\  1--13, 2025.
\newblock \doi{10.18653/v1/2025.unlp-1.1}.

\bibitem[Korablyov et~al.(2025)]{frontiers-2025-ukrainian-nlp}
Maksym Korablyov et~al.
\newblock Tokenization efficiency of current foundational large language models
  for the ukrainian language.
\newblock \emph{Frontiers in Artificial Intelligence}, 8, 2025.
\newblock \doi{10.3389/frai.2025.1538165}.

\bibitem[Land \& Bartolo(2024)Land and Bartolo]{land-bartolo-2024-fishing}
Sander Land and Max Bartolo.
\newblock Fishing for magikarp: Automatically detecting under-trained tokens in
  large language models.
\newblock \emph{arXiv preprint arXiv:2405.05417}, 2024.

\bibitem[{lang-uk}(2016)]{tokenize-uk-langek}
{lang-uk}.
\newblock tokenize-uk: Rule-based ukrainian word and sentence tokenizer.
\newblock \url{https://github.com/lang-uk/tokenize-uk}, 2016.

\bibitem[{Lapa LLM}(2025)]{lapa-tokenizer-release-2025}
{Lapa LLM}.
\newblock Lapa {U}krainian tokenizer.
\newblock \url{https://huggingface.co/lapa-llm/tokenizer}, 2025.

\bibitem[Liang et~al.(2023)Liang, Gonen, Mao, Hou, Goyal, Ghazvininejad,
  Zettlemoyer, and Khabsa]{liang-etal-2023-xlm}
Davis Liang, Hila Gonen, Yuning Mao, Rui Hou, Naman Goyal, Marjan
  Ghazvininejad, Luke Zettlemoyer, and Madian Khabsa.
\newblock {XLM-V}: Overcoming the vocabulary bottleneck in multilingual masked
  language models.
\newblock In \emph{Proceedings of EMNLP}, 2023.

\bibitem[Lotz et~al.(2025)Lotz, Lopes, Peitz, Setiawan, and
  Emili]{lotz-etal-2025-beyond}
Jonas~F. Lotz, Antonio~V. Lopes, Stephan Peitz, Hendra Setiawan, and Leonardo
  Emili.
\newblock Beyond text compression: Evaluating tokenizers across scales.
\newblock In \emph{Proceedings of ACL}, pp.\  32155--32173, 2025.

\bibitem[{NVIDIA}(2024)]{nvidia2024nemotron}
{NVIDIA}.
\newblock Nemotron-3-nano: A family of compact language models.
\newblock Technical report, NVIDIA, 2024.

\bibitem[Ociepa et~al.(2025)Ociepa, Flis, Wr{\'o}bel, Gwo{\'z}dziej, and
  Kinas]{ociepa-etal-2025-bielik}
Krzysztof Ociepa, {\L}ukasz Flis, Krzysztof Wr{\'o}bel, Adrian Gwo{\'z}dziej,
  and Remigiusz Kinas.
\newblock {B}ielik 7{B} v0.1: A {P}olish language model---development,
  insights, and evaluation.
\newblock \emph{Computer Science}, 26\penalty0 (4):\penalty0 131--161, 2025.
\newblock \doi{10.7494/csci.2025.26.4.7689}.

\bibitem[{OpenAI}(2025)]{openai-2025-gpt-oss}
{OpenAI}.
\newblock gpt-oss-120b \& gpt-oss-20b model card, 2025.

\bibitem[Paniv et~al.(2026)Paniv, Didenko, Haltiuk, Humennyy, Kravchenko,
  Kyslyi, Makovska, Orlovskyi, Ruban, Rudko, Senyk, Drushchak, Chaplynskyi, and
  Romanyshyn]{paniv-etal-2026-data}
Yurii Paniv, Bohdan Didenko, Mykola Haltiuk, Vladyslav Humennyy, Andrian
  Kravchenko, Roman Kyslyi, Viktoriia Makovska, Artem Orlovskyi, Bohdan Ruban,
  Maksym-Yurii Rudko, Anastasiia Senyk, Nazarii Drushchak, Dmytro Chaplynskyi,
  and Mariana Romanyshyn.
\newblock Data-efficient adaptation of multilingual {LLM}s to {U}krainian.
\newblock In Mariana Romanyshyn (ed.), \emph{Proceedings of the Fifth
  {U}krainian Natural Language Processing Conference ({UNLP} 2026)}, pp.\
  155--168, Lviv, Ukraine, May 2026. Association for Computational Linguistics.
\newblock ISBN 979-8-89176-359-3.
\newblock URL \url{https://aclanthology.org/2026.unlp-1.14/}.

\bibitem[Patil et~al.(2022)Patil, Talukdar, and
  Sarawagi]{patil-etal-2022-overlap}
Vaidehi Patil, Partha Talukdar, and Sunita Sarawagi.
\newblock Overlap-based vocabulary generation improves cross-lingual transfer
  among related languages.
\newblock In \emph{Proceedings of ACL}, pp.\  219--233, 2022.
\newblock \doi{10.18653/v1/2022.acl-long.18}.

\bibitem[Petrov et~al.(2024)Petrov, La~Malfa, Torr, and
  Bibi]{petrov-etal-2024-language}
Aleksandar Petrov, Emanuele La~Malfa, Philip H.~S. Torr, and Adel Bibi.
\newblock Language model tokenizers introduce unfairness between languages.
\newblock \emph{Advances in Neural Information Processing Systems}, 2024.

\bibitem[Purason et~al.(2025)Purason, Chizhov, Yamshchikov, and
  Fishel]{purason-etal-2025-teaching}
Taido Purason, Pavel Chizhov, Ivan~P. Yamshchikov, and Mark Fishel.
\newblock Teaching old tokenizers new words: Efficient tokenizer adaptation for
  pre-trained models.
\newblock \emph{arXiv preprint arXiv:2512.03989}, 2025.
\newblock URL \url{https://arxiv.org/abs/2512.03989}.

\bibitem[Radford et~al.(2019)Radford, Wu, Child, Luan, Amodei, and
  Sutskever]{radford2019language}
Alec Radford, Jeffrey Wu, Rewon Child, David Luan, Dario Amodei, and Ilya
  Sutskever.
\newblock Language models are unsupervised multitask learners.
\newblock Technical report, OpenAI, 2019.

\bibitem[Remy et~al.(2023)Remy, Delobelle, Berendt, Demuynck, and
  Demeester]{remy-etal-2023-tiktotok}
Fran{\c{c}}ois Remy, Pieter Delobelle, Bettina Berendt, Kris Demuynck, and
  Thomas Demeester.
\newblock Tik-to-tok: Translating language models one token at a time.
\newblock \emph{arXiv preprint arXiv:2310.03477}, 2023.

\bibitem[Rust et~al.(2021)Rust, Pfeiffer, Vuli{\'c}, Ruder, and
  Gurevych]{rust-etal-2021-good}
Phillip Rust, Jonas Pfeiffer, Ivan Vuli{\'c}, Sebastian Ruder, and Iryna
  Gurevych.
\newblock How good is your tokenizer? on the monolingual performance of
  multilingual language models.
\newblock In \emph{Proceedings of ACL-IJCNLP}, 2021.
\newblock \doi{10.18653/v1/2021.acl-long.243}.

\bibitem[Sharthak et~al.(2025)Sharthak, Pahalwan, Kamath, and
  Shirawalmath]{sharthak-etal-2025-tokenadapt}
Shaurya Sharthak, Vinayak Pahalwan, Adithya Kamath, and Adarsh Shirawalmath.
\newblock Achieving tokenizer flexibility in language models through heuristic
  adaptation and supertoken learning.
\newblock \emph{arXiv preprint arXiv:2505.09738}, 2025.
\newblock URL \url{https://arxiv.org/abs/2505.09738}.

\bibitem[Wang et~al.(2020)Wang, Mayhew, and Roth]{wang-etal-2020-extending}
Zihan Wang, Stephen Mayhew, and Dan Roth.
\newblock Extending multilingual {BERT} to low-resource languages.
\newblock In \emph{Findings of EMNLP}, pp.\  2649--2656, 2020.
\newblock \doi{10.18653/v1/2020.findings-emnlp.119}.

\bibitem[Wegmann et~al.(2025)Wegmann, Nguyen, and
  Jurgens]{wegmann-etal-2025-tokenization}
Anna Wegmann, Dong Nguyen, and David Jurgens.
\newblock Tokenization is sensitive to language variation.
\newblock In \emph{Findings of ACL}, pp.\  10958--10983, 2025.

\bibitem[Yang et~al.(2025)Yang, Li, Yang, Zhang, Hui, Zheng, Yu,
  et~al.]{yang-etal-2025-qwen3}
An~Yang, Anfeng Li, Baosong Yang, Beichen Zhang, Binyuan Hui, Bo~Zheng, Bowen
  Yu, et~al.
\newblock Qwen3 technical report.
\newblock \emph{arXiv preprint}, 2025.

\end{thebibliography}

\appendix
\section*{Appendix}

This appendix provides additional byte-level examples of the merge ordering problem, detailed removal statistics, per-script replacement counts, and summary statistics for the Ukrainian tokenizer builds.

\section{Additional Byte-Level Examples}
\label{app:byte-examples}

Words can route through even harsher byte-level structure than the {\fontencoding{T2A}\selectfont країн} example in \S\ref{sec:ordering}: in the original GPT-OSS tokenizer, ``{\fontencoding{T2A}\selectfont обґрунтування}'' tokenizes into byte spans \texttt{[D0 BE D0 B1]}, \texttt{[D2]}, \texttt{[91]}, \texttt{[D1 80 D1 83 D0 BD]}, \texttt{[D1 82 D1 83]}, and \texttt{[D0 B2 D0 B0 D0 BD D0 BD D1 8F]}. Thus the single character {\fontencoding{T2A}\selectfont ґ} (UTF-8 \texttt{D2 91}) is realized via two separate raw-byte tokens.
In such cases, visually natural character boundaries are irrelevant; what matters is the boundary induced by the current merge graph.

\section{Reachability Audit Details}
\label{app:stress-tests}

Table~\ref{tab:reachability} gives the complete pinned audit behind the verification claims in \S\ref{sec:experiments}.
Each result is reported as failures/tested items.
``Merge graph'' calls the Rust BPE model directly on serialized model keys through ordinary rank-ordered merging; ``full pipeline'' calls the tokenizer on decoded raw surfaces and also requires exact encode--decode roundtrips.
The guided-build surface row covers transferred donor surfaces, whereas the direct-transfer surface rows test every decodable inserted-node surface as an isolated input.

\begin{table}[h]
\centering
\scriptsize
\resizebox{\columnwidth}{!}{%
\begin{tabular}{@{}llcc@{}}
\toprule
\textbf{Audit scope} & \textbf{Execution path} & \textbf{Nemotron} & \textbf{GPT-OSS} \\
\midrule
\multicolumn{4}{@{}l}{\textit{BPE-guided builds}} \\
Full model-vocabulary entries & merge graph & 0 / 130,072 & 0 / 199,998 \\
Inserted BPE nodes & merge graph & 0 / 28,134 & 0 / 45,398 \\
Transferred donor surfaces & full pipeline: ordinary merge path & 0 / 27,975 & 0 / 45,232 \\
Retained same-ID rows & newly broken in merge graph & 0 / 101,938 & 0 / 154,600 \\
\midrule
\multicolumn{4}{@{}l}{\textit{Direct merge transfer from a target-native donor}} \\
Inserted BPE nodes & merge graph & 4,513 / 28,134 & 104 / 45,398 \\
Decodable isolated inserted surfaces & full pipeline: ordinary merge path & 4,512 / 28,126 & 104 / 45,387 \\
Decodable isolated inserted surfaces & full pipeline: whole-piece lookup & 0 / 28,126 & 0 / 45,387 \\
\bottomrule
\end{tabular}}
\caption{Reachability failures/tested items under Hugging Face Tokenizers 0.22.1. The upper block audits the BPE-guided builds. The lower block is an auxiliary runtime diagnostic using a byte-level BPE donor trained on the Ukrainian HPLT slice in each target's tokenizer family, not a matched-inventory compression baseline: whole-piece lookup emits valid isolated entries without repairing the merge graph.}
\label{tab:reachability}
\end{table}

Nemotron's full model-vocabulary denominator excludes 1,000 contents registered through the added-token layer; these pass a separate full-pipeline audit (0/1,000 failures).
The guided inserted-node totals exceed the donor-surface totals by 159/166 because they additionally include 152/153 target-script base nodes and 7/13 intermediate prerequisites.
The direct-transfer surface audit excludes 8/11 inserted nodes that do not decode to valid UTF-8 raw surfaces.
These distinct scopes therefore cannot be collapsed into a single ``unreachable tokens'' total.
Dependency-safe selection restored 2 optional Nemotron prerequisites and 14 GPT-OSS prerequisites, replacing each with the next eligible script-selected row while keeping the slot budgets exact.
The iterative audit terminated with no newly broken retained entries; neither target required adding a mandatory-reset dependent to the removal closure.

\section{Token Removal Details}
\label{app:removal}

Table~\ref{tab:ua_removal} separates the target-script reset from the script-selected fill rows.

\begin{table}[h]
\centering
\small
\resizebox{\columnwidth}{!}{%
\begin{tabular}{@{}lrr@{}}
\toprule
\textbf{Selected category} & \textbf{Nemotron} & \textbf{GPT-OSS} \\
\midrule
Cyrillic target reset & 7,686 & 14,214 \\
Other selected writing systems & 20,448 & 31,184 \\
\midrule
\textbf{Total selected rows} & \textbf{28,134} & \textbf{45,398} \\
\bottomrule
\end{tabular}}
\caption{Final replacement-row budgets. The target reset takes precedence over the configured selection filter; the other rows are selected under the writing-system policy and dependency repair described in \S\ref{sec:method}.}
\label{tab:ua_removal}
\end{table}

\section{Replaced Tokens by Writing System}
\label{app:replaced}

Table~\ref{tab:script_selected} gives the final manifest-derived counts for every writing system selected in either target.

\begin{table}[h]
\centering
\small
\begin{tabular}{@{}lr@{\hspace{2em}}r@{}}
\toprule
\textbf{Writing system} & \textbf{Nemotron} & \textbf{GPT-OSS} \\
\midrule
Arabic & 7,069 & 6,599 \\
Armenian & 811 & 1,396 \\
Bengali & 614 & 1,745 \\
Cyrillic & 7,686 & 14,214 \\
Devanagari & 1,137 & 3,218 \\
Georgian & 373 & 0 \\
Gujarati & 148 & 1,315 \\
Gurmukhi & 105 & 238 \\
Han & 2,800 & 5,960 \\
Hangul & 3,330 & 1,888 \\
Hebrew & 744 & 1,931 \\
Japanese (Hiragana/Katakana) & 962 & 559 \\
Kannada & 415 & 1,084 \\
Khmer & 0 & 286 \\
Lao & 0 & 2 \\
Malayalam & 288 & 1,359 \\
Myanmar & 164 & 229 \\
Oriya & 0 & 34 \\
Sinhala & 0 & 233 \\
Tamil & 400 & 789 \\
Telugu & 653 & 1,097 \\
Thai & 435 & 1,222 \\
\midrule
\textbf{Total} & \textbf{28,134} & \textbf{45,398} \\
\bottomrule
\end{tabular}
\caption{Selected replacement rows by primary Unicode writing-system classification after dependency repair. This is a construction breakdown, not an evaluation of language preservation; zero denotes a writing system not selected for that target.}
\label{tab:script_selected}
\end{table}

\FloatBarrier

\section{Extended Ukrainian Evaluation Metrics}
\label{app:extended_metrics}

Table~\ref{tab:extended_metrics} reports exact token counts and fertility for the unmodified bases and our final tokenizers on every evaluated corpus.

\begin{table*}[t]
\centering
\scriptsize
\setlength{\tabcolsep}{4pt}
\resizebox{\textwidth}{!}{
\begin{tabular}{@{}lcccc@{}}
\toprule
\textbf{Eval set} & \textbf{Nemotron base} & \textbf{Nemotron ours} & \textbf{GPT-OSS base} & \textbf{GPT-OSS ours} \\
\midrule
\texttt{lang-uk/malyuk} [100k] &
\shortstack{62,087,149\\2.711} &
\shortstack{41,303,450\\1.804} &
\shortstack{59,447,036\\2.596} &
\shortstack{37,686,906\\1.646} \\

\texttt{allenai/c4(en)} [100k] &
\shortstack{47,630,139\\1.317} &
\shortstack{47,646,064\\1.317} &
\shortstack{45,423,925\\1.256} &
\shortstack{45,445,165\\1.256} \\

\texttt{allenai/c4(es,fr,it,de)} [100k each] &
\shortstack{365,218,644\\1.843} &
\shortstack{365,264,894\\1.843} &
\shortstack{335,188,687\\1.691} &
\shortstack{335,247,179\\1.692} \\

\texttt{QIRIM/crh\_monocorpus} [94] &
\shortstack{6,623,516\\3.545} &
\shortstack{6,639,984\\3.554} &
\shortstack{5,995,822\\3.209} &
\shortstack{6,192,238\\3.314} \\

\texttt{allenai/c4(ru)} [100k] &
\shortstack{107,233,038\\2.520} &
\shortstack{113,120,042\\2.658} &
\shortstack{91,824,464\\2.158} &
\shortstack{101,077,477\\2.375} \\

\texttt{allenai/c4(bg)} [100k] &
\shortstack{108,691,963\\2.436} &
\shortstack{117,815,178\\2.640} &
\shortstack{102,472,523\\2.296} &
\shortstack{108,584,543\\2.433} \\

\texttt{allenai/c4(be)} [100k] &
\shortstack{135,489,439\\3.140} &
\shortstack{143,859,938\\3.334} &
\shortstack{119,587,038\\2.771} &
\shortstack{135,796,072\\3.147} \\
\bottomrule
\end{tabular}}
\caption{Pinned tokenizer-level evaluation. Each cell reports \textit{tokens / tokens per whitespace-delimited word}; EU is shown as one micro-aggregated row after evaluating 100k examples in each language.}
\label{tab:extended_metrics}
\end{table*}

\FloatBarrier

\section{Ukrainian Tokenizer Statistics}
\label{app:ua_stats}

Table~\ref{tab:ua_stats} summarizes the surgery and validation counts for the ID-preserving Ukrainian tokenizers.

\begin{table}[h]
\centering
\small
\resizebox{\columnwidth}{!}{%
\begin{tabular}{@{}lrr@{}}
\toprule
\textbf{Metric} & \textbf{Nemotron} & \textbf{GPT-OSS} \\
\midrule
Base model & Nemotron-3 & GPT-OSS-20B \\
Base model-vocabulary size & 131,072 & 199,998 \\
Selected replacement rows & 28,134 & 45,398 \\
Inserted donor surfaces & 27,975 & 45,232 \\
Inserted target-script base nodes & 152 & 153 \\
Inserted intermediate prerequisite nodes & 7 & 13 \\
Total inserted BPE nodes & 28,134 & 45,398 \\
Strict-broken inserted nodes & 0 & 0 \\
Operationally broken donor surfaces (lookup disabled) & 0 & 0 \\
Dependency candidates restored/replaced & 2 & 14 \\
Newly strict-broken retained rows & 0 & 0 \\
Strict-broken full model-vocabulary entries & 0 & 0 \\
Final model-vocabulary size & 131,072 & 199,998 \\
\bottomrule
\end{tabular}}
\caption{Tokenizer surgery statistics. ``Strict'' and ``operational'' use the protocol definitions in \S\ref{sec:ordering} and \S\ref{sec:runtime-lookup}; the full Nemotron model-vocabulary audit excludes 1,000 contents registered through the added-token layer and checks them operationally.}
\label{tab:ua_stats}
\end{table}

\end{document}